\documentclass[10pt, a4paper]{article}
\usepackage{titlesec}
\usepackage[belowskip=0pt,aboveskip=0pt]{caption}
\usepackage{lrec2026}
\usepackage{booktabs}
\usepackage{graphicx}
\usepackage{pdflscape}
\usepackage{url}
\usepackage{titlesec}
\titlespacing*{\section}{0pt}{1ex}{0.4ex}
\titlespacing*{\subsection}{0pt}{1ex}{0.4ex}
\titlespacing*{\paragraph}{0pt}{1ex}{0.4ex}

\title{AfriVoices-KE: A Multilingual Speech Dataset for Kenyan Languages}

\name{\parbox{\linewidth}{\centering \large
      Lilian Wanzare$^{1,*}$, Cynthia Amol$^{1}$, Ezekiel Maina$^{1}$, Nelson Odhiambo$^{1}$, \\
      Hope Kerubo$^{1}$, Leila Misula$^{1}$, Vivian Oloo$^{1}$, Rennish Mboya$^{1}$, \\
      Edwin Onkoba$^{1}$, Edward Ombui$^{2}$, Joseph Muguro$^{3}$, Ciira wa Maina$^{3}$, \\
      Andrew Kipkebut$^{4}$, Alfred Omondi Otom$^{5,6}$, Ian Ndung'u Kang'ethe$^{7}$, \\
      Angela Wambui Kanyi$^{7}$, Brian Gichana Omwenga$^{7}$ 
      \vspace{0.2cm} 
      }}

\address{\parbox{\linewidth}{\centering \normalsize 
         $^{1}$Maseno University, Kenya \\
         $^{2}$United States International University, Kenya \\
         $^{3}$Dedan Kimathi University of Technology, Kenya \\
         $^{4}$Kabarak University, Kenya \\
         $^{5}$Jaramogi Oginga Odinga University of Science and Technology, Kenya \\
         $^{6}$KU Leuven, Belgium \\
         $^{7}$Tech Innovators Network (THiNK), Kenya \\ \vspace{0.2cm}
         $^{*}$ldwanzare@maseno.ac.ke
         \vspace{0.4cm} 
         }}
         
\abstract{
AfriVoices-KE is a large-scale multilingual speech dataset comprising approximately 3,000 hours 
of audio across five Kenyan languages: Dholuo, Kikuyu, Kalenjin, Maasai, and Somali. The dataset 
includes 750 hours of scripted speech and 2,250 hours of spontaneous speech, collected from 4,777 
native speakers across diverse regions and demographics. This work addresses the critical 
underrepresentation of African languages in speech technology by providing a high-quality, 
linguistically diverse resource. Data collection followed a dual methodology: scripted recordings 
drew from compiled text corpora, translations, and domain-specific generated sentences spanning 
eleven domains relevant to the Kenyan context, while unscripted speech was elicited through 
textual and image prompts to capture natural linguistic variation and dialectal nuances. A 
customized mobile application enabled contributors to record using smartphones. Quality assurance 
operated at multiple layers, encompassing automated signal-to-noise ratio validation prior to 
recording and human review for content accuracy. Though the project encountered challenges 
common to low-resource settings, including unreliable infrastructure, device compatibility 
issues, and community trust barriers, these were mitigated through local mobilizers, stakeholder 
partnerships, and adaptive training protocols. AfriVoices-KE provides a foundational resource for 
developing inclusive automatic speech recognition and text-to-speech systems, while advancing the 
digital preservation of Kenya's linguistic heritage.
\\ \newline }

\begin{document}

\maketitleabstract

\section{Introduction}

The rapid advancement of speech technologies, such as Automatic Speech Recognition (ASR) and 
Text-To-Speech (TTS) systems, has revolutionized human-computer interaction, enabling 
applications in healthcare, education, agriculture, and financial services. However, these 
technologies remain heavily skewed towards high-resource languages like English and other 
Indo-European languages, leaving low-resource languages, particularly those in Africa, 
severely underrepresented in speech datasets \citep{emezue2025naijavoices}.

Kenya's linguistic environment includes over 60 languages according to Ethnologue 
\citep{eberhard2025ethnologue}, encompassing Bantu, Nilotic, and Cushitic language families. 
Indigenous languages such as Dholuo, Kalenjin, Maasai, Kikuyu, and Kenya Somali face 
significant data scarcity. The distribution of digital resources is highly uneven: Swahili is 
the most documented language, with over 20 available datasets covering both text and speech. In 
contrast, over 70\% of publicly documented datasets for Kenyan languages are text-only, 
primarily supporting tasks like Machine Translation (MT) \citep{amol2024state}. As a result, 
ASR development for indigenous Kenyan languages is severely limited, and more than 50 languages 
have no publicly available digital resources at all \citep{amol2024state}. This imbalance presents a core challenge 
for digital inclusion, restricting access to voice-enabled services such as mobile health 
advisories, agricultural information systems, and digital government platforms, which are 
critical in a country where literacy rates vary and oral traditions are central to communication.

The AfriVoices-KE project aimed to fill gaps in low-resource languages by creating a large-scale, multilingual speech dataset targeting approximately 3,000 hours of recordings 
across five Kenyan languages: Dholuo, Kalenjin, Maasai, Kikuyu, and Kenya Somali. The project targeted between 500 and 750 hours per language: Dholuo (750 hrs), Gikuyu (750 hrs), Somali 
(500 hrs), Kalenjin, covering Nandi and Kipsigis dialects, (500 hrs), and Maasai (500 hrs). 
All datasets comprised a mix of scripted and unscripted speech, approximately 75\% unscripted and 25\% scripted (see Section ~\ref{sec:mode}. The dataset balanced scripted readings with unscripted spontaneous speech, elicited through diverse prompts across different domains relevant to the Kenyan context, as detailed in Section~\ref{sec:domains}.

The study pursued the following objectives:
\begin{enumerate}
 \setlength{\itemsep}{2pt}
    \setlength{\parskip}{0pt}
    \setlength{\topsep}{2pt}
    \item Compile a comprehensive text corpus for scripted recordings by collating, translating, 
    and generating sentences in the target languages. 
    \item Develop diverse prompts (text, visual, audio) to elicit spontaneous speech, capturing 
    natural language use and dialectal variations.
    \item Conduct large-scale data collection using a customized mobile app.
    \item Transcribe all recordings with high accuracy, annotating code-switched terms and 
    preserving dialectal nuances to enhance dataset utility.
\end{enumerate}

By leveraging a crowd-sourced mobile app, rigorous speech quality assessment, and native-speaker 
transcription with code-switching annotations, AfriVoices-KE surpasses existing Kenyan datasets 
in volume, diversity, and methodological rigor. It addresses dialectal variations and ethical 
community engagement, contributing to the digital preservation of Kenyan linguistic heritage and 
enabling inclusive AI for under-served communities.

The remainder of this paper is organized as follows. Section~\ref{sec:related} reviews related work across the continent, while Section~\ref{sec:preliminary} discusses the preliminary decisions that guided the project design. Section~\ref{sec:methodology} and Section~\ref{sec:stats} describe the data curation methodology and provide a detailed overview of the collected data, respectively. Finally, Sections~\ref{sec:ethics} and~\ref{sec:release} address the ethical considerations of the project and outline the procedures for data release.

\section{Related Work}
\label{sec:related} 
With over 2,000 languages spoken across the continent \citep{eberhard2025ethnologue}, African 
linguistic diversity is vast, yet the lack of robust datasets limits the development of inclusive 
language technology tools. Data used for training most language models are often sourced from web-crawled sources 
\citep{penedo2023refinedweb} such as CommonCrawl, which are significantly less 
resource-intensive compared to collection from primary data sources. In the case of low-resource 
languages whose communities have a limited digital presence \citep{nigatu2024digital}, these web 
sources are not ideal for producing inclusive, representative samples. Furthermore, some openly 
available multilingual speech datasets for African languages, such as Bible TTS 
\citep{meyer2022bibletts}, are skewed towards a single domain, religion, limiting their 
utility for the diverse, community-facing use cases increasingly demanded by the adoption of AI 
and automation across developing countries in Africa.

Efforts to address this gap have made notable strides, particularly for Kenyan languages. The 
Kencorpus dataset provides 177 hours of speech across Swahili, Dholuo, and Luhya dialects, 
collected from community sources \citep{wanjawa2023kencorpus}. A corpus covering Kidawida, 
Kalenjin, and Dholuo provides approximately 268 hours of speech gathered through crowd-sourced 
conversations and text readings in domains such as public health and agriculture 
\citep{mbogho2025building}. MakerereNLP aimed to develop open speech datasets for East African 
languages from Kenya, Uganda, and Tanzania, though specific dataset sizes remain in progress 
\citep{meyer2022bibletts}, and smaller resources such as Bible TTS provide limited speech for 
Kikuyu and Dholuo, often restricted to religious texts \citep{olatunji2023afrispeech}.

Beyond Kenya, substantial ASR corpora have been developed across other African and comparable 
low-resource language communities. In West Africa, the NaijaVoices dataset provides 1,800 hours 
of authentic speech across Igbo, Hausa, and Yor\`{u}b\'{a} \citep{emezue2025naijavoices}, while 
\`{I}ro\`{y}\`{i}nSpeech offers approximately 48 hours of curated Yor\`{u}b\'{a} speech from 
news and creative writing domains \citep{ogunremi2024iroyinspeech}, and SautiDB-Naija addresses 
Nigerian-accented English for accent benchmarking \citep{Afonja2021}. In southern Africa, the 
BembaSpeech corpus provides over 24 hours of read speech for the Bemba language of Zambia 
\citep{Sikasote2022}. For Semitic and Cushitic languages, King-ASR-749 provides 233.6 hours of 
mobile-recorded Amharic speech \citep{Emiru2020}, while the pan-African AfriSpeech-200 corpus 
contributes 200 hours of accented English across clinical and general domains from 120 
accents across 13 African countries, including Swahili-influenced variants 
\citep{olatunji2023afrispeech}, and Kinyarwanda has amassed over 1,183 validated hours through 
the Common Voice platform. Global ASR resources frequently used for cross-lingual transfer 
include the FLEURS benchmark, covering 102 languages with roughly 12 hours of supervised speech 
per language \citep{Conneau2022}, and the LibriSpeech corpus, a large read-speech resource 
derived from audiobooks in English, German, and French \citep{Panayotov2020}.

\section{Preliminary guiding decisions}
\label{sec:preliminary}
As highlighted by \citet{Okorie2024}, careful and systematic planning, including ethical considerations, are essential in the design of any data collection initiative. Four critical decisions were identified as foundational to the process: the \textit{selection of languages} to be included in the data collection, the appropriate \textit{mode for voice data collection}, the most suitable \textit{technological platform} to support the different data collection modes and finally, the \textit{domains to be represented in the corpus}. This section highlights the choices we made in the project.

\subsection{Languages}
\label{sec:languages}
\begin{figure}[!ht]
\begin{center}
\includegraphics[trim={0 0 0 2.7cm},clip,width=\columnwidth]{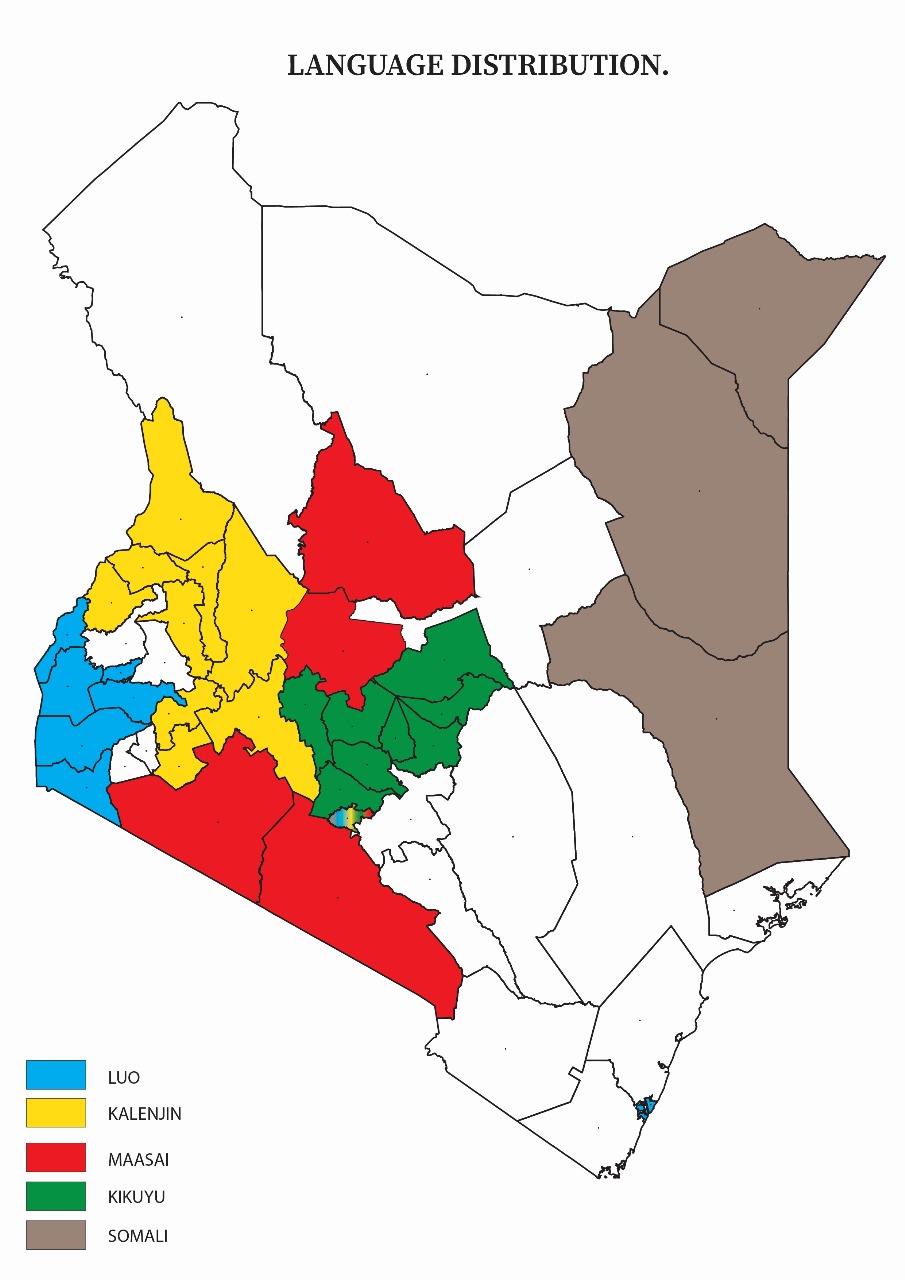}
\caption{Geographical distribution of the five Kenyan languages covered in AfriVoices-KE.}
\label{fig:geog}
\end{center}
\vspace*{-2em}
\end{figure}

In the data collection, we needed to ensure that every language family in Kenya
is represented. We also needed to put into consideration the number of speakers for the different
languages so as to capture as many language speakers as possible. 
Five languages were selected to ensure representation across 
Kenya's major language families, while also maximising speaker coverage. The selected languages 
span the Nilotic family,  represented by Dholuo (Lake Nilotic), Kalenjin (Highland Nilotic), 
and Maasai (Plains Nilotic), alongside the Cushitic family (Somali) and the Bantu family 
(Kikuyu). Figure~\ref{fig:geog} illustrates the geographical distribution of these languages 
across Kenya.

\textbf{Dholuo} (ISO 639-3: \textit{luo}) is a River-Lake Nilotic language spoken by over 4.2 
million people in western Kenya, with additional speakers in northern Tanzania 
\citep{ldc2020dholuo}. The language exhibits two principal dialects, Milambo (southern) and 
Nyandwat (northern), reflecting geographical variation across counties such as Kisumu, Homa 
Bay, Migori, and Siaya \citep{owego2025dialects}.

\textbf{Kikuyu} (ISO 639-3: \textit{kik}) is a Bantu language and the largest ethnic language community in Kenya, with approximately 8.15 million speakers concentrated in the central region. 
This study adopts a five-dialect framework, covering Kiambu, Murang'a, Nyeri, and Kirinyaga 
(subdivided into K\~{i}-Ndia and G\~{i}-G\~{i}ch\~{u}g\~{u}), to capture regional 
phonological and sociolinguistic variation \citep{eberhard2025ethnologue}.

\textbf{Kalenjin} (ISO 639-3: \textit{kln}) is a Southern Nilotic language cluster spoken by 
approximately 6.3 million people in Kenya's Rift Valley. Data collection focused on the two 
most widely spoken dialect groupings, Nandi (Nandi and Uasin Gishu counties) and Kipsigis 
(Kericho and Bomet counties), which together account for an estimated 60 - 70\% of all Kalenjin 
speakers \citep{eberhard2025ethnologue}.

\textbf{Maasai} (ISO 639-3: \textit{mas}) is a Plains Nilotic language, known as Maa, spoken 
by over 2.1 million people across Kenya's South Rift region and northern Tanzania. The language 
is divided into North Maa and South Maa dialect clusters; this study focuses on South Maa 
varieties, which exhibit approximately 80 - 90\% mutual intelligibility 
\citep{eberhard2025ethnologue}.

\textbf{Somali} (ISO 639-3: \textit{som}) is an Afro-Asiatic Cushitic language spoken by more 
than 25 million people across the Horn of Africa. In Kenya, it is predominantly spoken in 
Wajir, Garissa, and Mandera counties. This study collected data based on the Northern Somali 
(Maxatiri) dialect, which serves as the standard written variety \citep{eberhard2025ethnologue}.

\subsection{Choice of mode of collection}
\label{sec:mode}
Large-scale speech data collection typically adopts either scripted or unscripted approaches, each with distinct strengths and limitations. In the case of this project, scripted voice collection involved participants reading predetermined sentences, producing controlled short audio that enabled efficient and scalable dataset development; however, because it follows a text-to-audio process, it often lacks the spontaneity, prosodic variation, disfluencies, code-switching, and dialectal richness characteristic of natural speech, potentially limiting ecological validity and inclusivity in multilingual contexts. Unscripted voice collection, by contrast, prompts contributors to speak spontaneously on given topics, thereby capturing authentic phrasing, accents, environmental noise, and real-world speech phenomena that enhance the robustness and contextual adaptability of automatic speech recognition (ASR) systems. Although unscripted data introduce technical challenges, including variable recording conditions and segmentation constraints, they provide a stronger foundation for developing inclusive and resilient speech technologies. In this project, a deliberate balance was adopted, with approximately 75\% of the collected data being unscripted and 25\% scripted, combining scalability with ecological validity.

\subsection{Choice of data collection platform}
The  architectural design and functional capacity of the data collection platform was a key consideration. Four principal requirements guided platform selection and development. First, the system was required to support both scripted and unscripted voice collection modes and to operate efficiently across devices, particularly mobile phones. Second, the platform needed to enable linguistic and regional customization, including expanded local metadata fields,such as dialect, level of education, county, and constituency, alongside standard demographic variables (e.g., age and gender). Third, integrated workflows for transcription and validation were necessary to ensure data quality, particularly for unscripted recordings. Fourth, given that contributors were remunerated, the platform required robust mechanisms for tracking participation metrics and output volumes to ensure transparency and accountability. Although existing initiatives such as Karya\footnote{\url{https://www.karya.in/}}, Digital Umuganda\footnote{\url{https://digitalumuganda.com/}}, and Mozilla Common Voice\footnote{\url{https://commonvoice.mozilla.org/}} offer established models for speech data collection, limitations related to open-source accessibility, customization capacity, and resource constraints rendered their adaptation impractical. Consequently, a dedicated open-source mobile application was developed to address the specific technical, linguistic, and operational requirements of the project.
\subsection{Choice of domains}
\label{sec:domains}
Domain diversification ensures linguistic richness and enables models to perform across varied real-world applications. The dataset was organized around eleven domains: Agriculture \& Food, 
Everyday Scenarios, Financial Transactions, Digital Government Services, Named Entity 
Recognition (NER), Role Play Conversations, Extempore Stories, Healthcare, News \& Media, 
Education \& Technology, and Customer Care Scenarios. Agriculture \& Food was allocated 40\% of 
total data, reflecting its relevance as a downstream use case in Kenya and across Africa. 

\section{Dataset Curation}
\label{sec:methodology}
This section covers the process of curating the dataset from development of the tool, scripted and unscripted data collection and quality control, highlighting on the challenges and opportunities.  

\subsection{Data Collection Tool}
\label{sec:tool}
The Custom Voice Collection App was developed as a mobile-first, open-source platform supporting 
both scripted (Section~\ref{sec:scripted}) and unscripted (Section~\ref{sec:unscripted}) data 
collection. The system was built around four principles: accessibility, validation, cultural 
sensitivity, and offline resilience, and deployed across web and mobile interfaces to accommodate varying levels of connectivity and device access.
The system supported multiple roles: contributors who recorded audio clips, reviewers/validators, who assessed the quality of audio submissions, transcribers who transcribed the approved audio clips, super-reviewers, who provided an additional oversight layer and administrators, who managed user participation and platform health.
The mobile application allowed contributors to register, record, replay, and submit audio, with personal dashboards displaying the number of clips submitted, accepted, rejected, and pending, a feature that contributors reported as a strong motivator for sustained engagement.   

\paragraph{Challenges.}
The most frequently reported challenge was the instability of the data collection application particularly in the early stages of deployment, slow loading times, frequent crashes during registration and login, and high data consumption.
Device compatibility was a significant barrier, as the application did not function reliably on older budget smartphone models, excluding a portion of potential contributors.  Poor network connectivity in rural areas further hindered audio uploads. 

The key technical challenges were platform scalability under concurrent load from over 12,000 
active users, cross-device compatibility across low-memory Android handsets, offline functionality and maintaining 
audio integrity from variable acoustic environments. These were addressed through a containerized 
microservices architecture with auto-scaling, an automated validation pipeline for signal to noise ratio and audio 
duration checks. Future initiatives would benefit from a more robust, offline-capable platform.

\subsection{Recruitment and on-boarding}
 Data collection process was coordinated by resource personnel with established connections to local communities. Resource persons were recruited through formal contracts, selected on the basis of multilingual proficiency in Swahili, English, and the target local language, alongside basic data management skills. 
 We recruited contributors primarily through a network of local mobilizers who were familiar with
the community demographics. These mobilizers, often respected community figures or leaders, were instrumental in identifying and inviting individuals who met the project’s demographic criteria.
 Demographic balance was enforced through quotas targeting 50\% male and female representation and at least 15\% contribution per age bracket (18--29, 30--39, 40--49, 50--59, 
60+), with both rural and urban contributors included across language-speaking counties targeting 60\% to 80\% of the contributors coming from the districts where the languages are predominantly spoken. Other demographic information included dialect, education and employment status of the contributors. 
Contributors were trained and on-boarded to understand the project goal and the data collection procedure. 
\paragraph{Challenges:}
Recruiting a demographically balanced cohort proved difficult, with persistent under-representation of individuals aged 50 and above and lower-than-targeted female participation in some areas. Building trust presented a notable challenge. A significant number of participants expressed discomfort with providing personal information like ID and phone numbers due to privacy concerns. There was also initial skepticism about the project’s intentions and
payment schedules with many participants expecting immediate cash disbursements upon completion of recordings. The fieldwork indicates that leveraging trusted local mobilizers and partnerships with government offices, setting clear expectations during on-boarding, and establishing post-training WhatsApp groups for ongoing support were critical strategies for building trust, addressing logistical challenges, and sustaining participant engagement.

\subsection{Scripted Speech Collection}
\label{sec:scripted}
\begin{figure}[!ht]
\begin{center}
\includegraphics[width=\columnwidth]{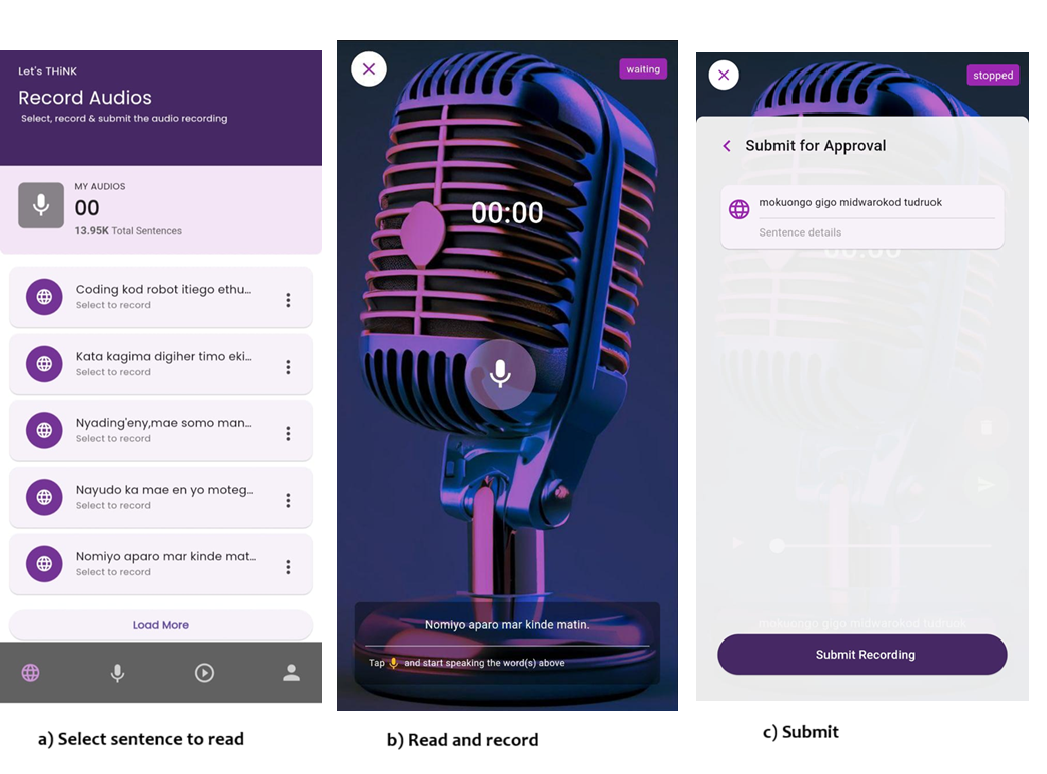}
\caption{The Custom Voice Collection App: scripted speech recording workflow, illustrating 
sentence selection, noise check, recording and playback, and submission confirmation.}
\label{fig:scripted_tool}
\end{center}
\vspace*{-2em}
\end{figure}

Scripted speech accounted for 25\% of the total dataset (750 hours), with contributors reading 
pre-prepared sentences across eleven domains. Source texts were compiled through three 
strategies: (1) aggregation from linguistic databases, academic repositories, and digital 
archives; (2) translation of English and Swahili texts by native speakers for languages with 
limited digital resources, such as Maasai and Somali; and (3) sentence generation for 
domain-specific and phonetically underrepresented content, including Named Entities and Extempore Stories.
All collected text included metadata such as source, dialect, and topic.  Sentence length and complexity were intentionally varied to
capture a wide phonetic range. All texts were cleaned, normalized, and segmented, with orthography and diacritics 
standardized per language. Pilot trials established that one hour of raw audio corresponds to approximately 350 read sentences, while one hour of validated speech requires 800 - 1,000 sentences.
The contributors were only required to choose a sentence then record themselves while reading it.  The scripted recording workflow is illustrated in Figure~\ref{fig:scripted_tool}.

\paragraph{Challenges.} Sourcing 
culturally relevant domain-specific texts from fragmented repositories was difficult, especially 
for technical domains such as Healthcare and Digital Government Services. Translation consistency 
was also variable across contributors with differing levels of proficiency. These were addressed 
through recruitment quotas, supplementary sentence generation, dialect-specific glossaries, and 
a batch tracker system with structured validator feedback.

\subsection{Unscripted Speech Collection}
\label{sec:unscripted}

Unscripted speech comprised 75\% of the total dataset (2,250 hours), collected through 
open-ended textual and image-based prompts designed to elicit spontaneous monologues reflecting 
natural prosody, dialectal variation, disfluencies, and contextually rich utterances. A total of 4,727 textual prompts and 5,469 image prompts were developed across all eleven domains, with 
Agriculture \& Food constituting the largest share (2,624 text; 2,478 image prompts). Prompts were formulated by language leads (LLs) and resource persons (RPs) in English or Swahili with 
parallel translations into all five target languages. Each language had a team in charge of sourcing for culturally relevant images. Audio and video prompts (196 items) were collected but 
not fully integrated due to bandwidth and file-size constraints. 

\begin{figure}[!ht]
\begin{center}
\includegraphics[width=\columnwidth]{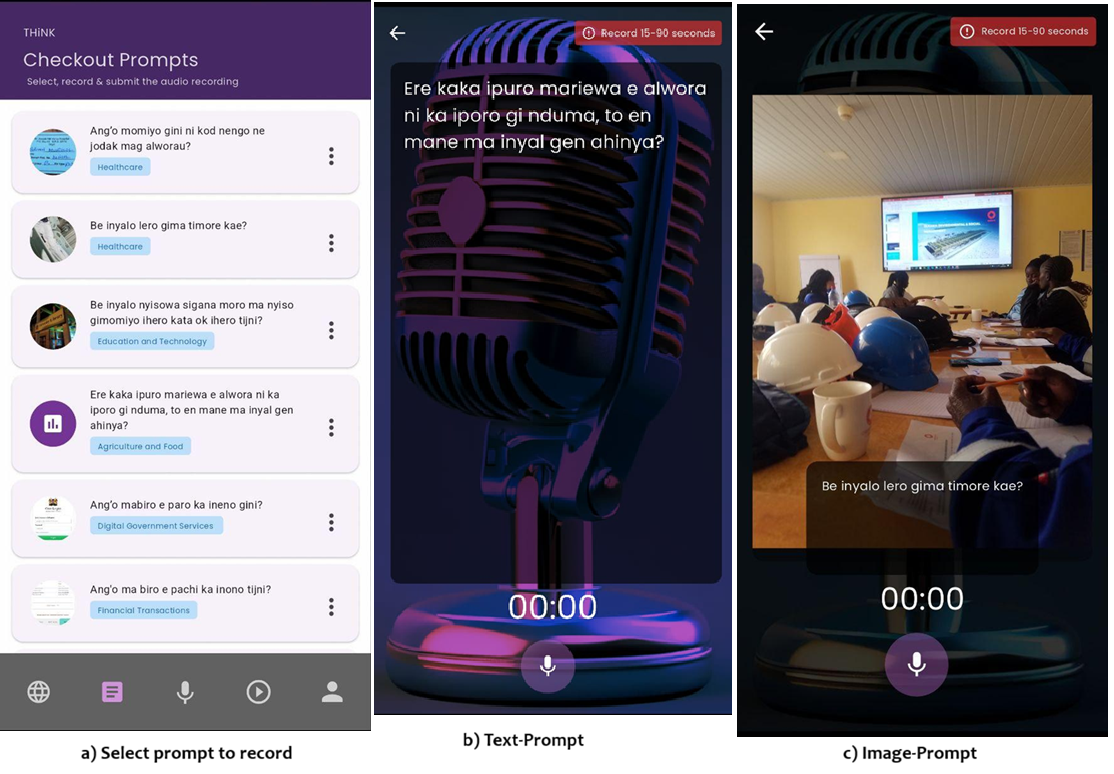}
\caption{The Custom Voice Collection App: unscripted speech recording workflow, illustrating 
prompt selection, noise check, recording and playback, and submission confirmation.}
\label{fig:unscripted_tool}
\end{center}
\vspace*{-2em}
\end{figure}

The contributors were to choose a prompt, then record themselves answering the question or talking about the image. Unscripted recordings were capped at 90 seconds per prompt response, with contributors advised to respond directly 
without repeating the prompt question or beginning with phrases such as ``In this photo\ldots''.  Participants could review and re-record before submission. 
The unscripted recording workflow is illustrated in Figure~\ref{fig:unscripted_tool}.

\begin{figure}[!ht]
\begin{center}
\includegraphics[width=\columnwidth]{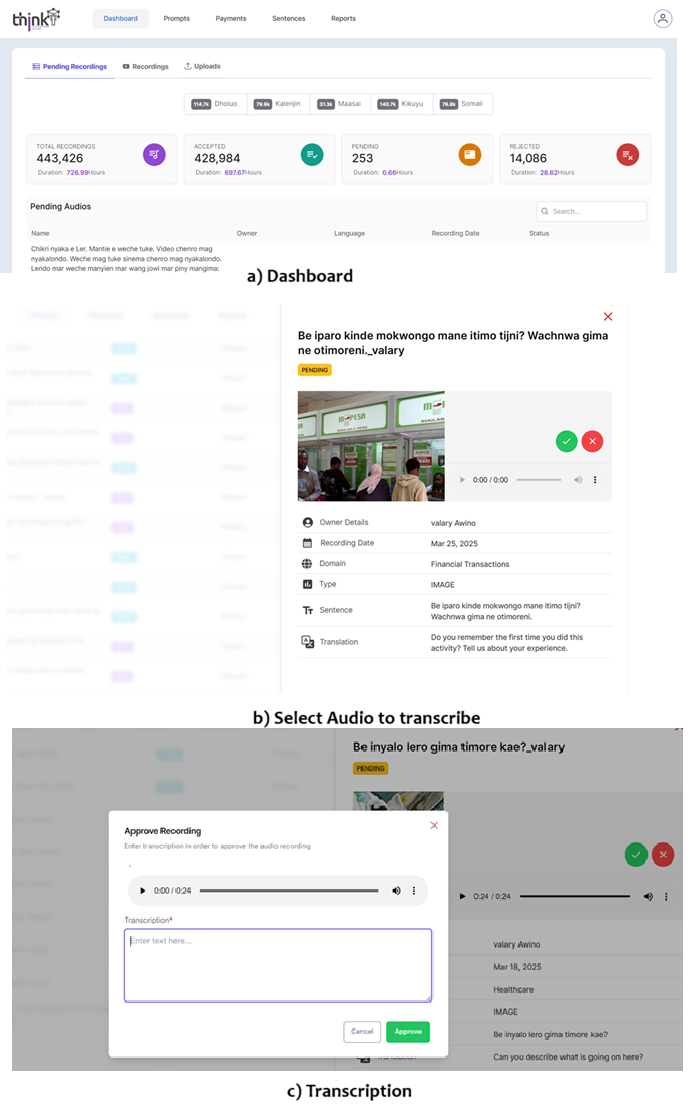}
\caption{Transcription workflow in the Custom App: from dashboard login and pending queue 
selection, through verbatim transcription and self-review, to final submission and dashboard 
update.}
\label{fig:transcription}
\end{center}
\vspace*{-2em}
\end{figure}

\paragraph{Transcription.} Unscripted audio was transcribed verbatim by trained native speakers 
using the Custom Transcription App (Figure~\ref{fig:transcription}). Transcribers followed 
standardized guidelines covering number formatting, diacritics, punctuation, and 
language-specific phonemic conventions. Code-switched words were tagged \texttt{[CS]...[CS]}, 
pauses longer than two seconds marked \texttt{[pause]}, and unclear speech annotated 
\texttt{[?]} (files with more than two such instances were rejected). A two-stage QA process 
was applied: Level~1 required the transcriber to reject audio containing multiple speakers, 
excessive background noise, or off-topic content before transcription began; Level~2 involved 
dedicated QA checkers reviewing completed transcripts for accuracy and adherence to guidelines. 

\paragraph{Challenges: } Several data quality challenges were observed during collection and transcription. Variability in recording conditions, including background noise, low audio volume, and overlapping speech, reduced clarity and increased transcription time. Additionally, frequent code-switching across languages and dialects, the presence of multiple speakers within single recordings, and natural speech phenomena such as pauses, repetitions, fillers, and incomplete utterances complicated speaker identification, verbatim transcription, and accurate preservation of meaning. 
Time and resource constraints posed significant challenges, as transcription, and review processes are inherently labor-intensive, requiring careful balancing of quality and project timelines. High-quality recordings substantially improved transcription accuracy and reduced review time, underscoring the importance of sound data collection practices. Furthermore, the recruitment of qualified language resource persons was critical to ensuring linguistic accuracy and cultural integrity, while the provision of clear guidelines enhanced efficiency, consistency, and collaborative workflow.

\subsection{Quality Control}
\label{sec:qc}

Quality assurance operated across multiple layers throughout the collection pipeline. At the 
point of recording, an automated noise detection module enforced minimum signal-to-noise ratio 
(SNR) requirements of $\sim$40~dB and frequency response validation at $\sim$48~kHz before a 
submission could be accepted. Contributors were trained during on-boarding to maintain quiet 
acoustic environments, speak audibly at a natural pace, avoid repeating the prompt question, and 
refrain from allowing proxy recordings, clips violating these criteria were subject to rejection 
during validation.

Submitted audio underwent human review by trained validators, who accepted or rejected recordings based on intelligibility, fluency, noise levels, and language compliance. There were between 5 - 15 validators for scripted and 20 - 40 validators for unscripted per language. A super-reviewer tier audited validator decisions for consistency, with structured feedback returned to contributors and decisions logged systematically. For each language, there were between 5 to 10 super-reviewers.  For unscripted data, the two-tier 
transcription QA process described in Section~\ref{sec:unscripted} ensured verbatim accuracy and dialect fidelity. Ten percent of all transcripts were double-verified by QA checkers and tagged as gold standard, providing a reliable benchmark subset for downstream model evaluation.

\section{Dataset Description}
\label{sec:stats}

The AfriVoices-KE dataset comprises about 3,000 hours of audio across five languages, collected from 
4,677 contributors. Scripted recordings account for 669 hours (22.3\%) and unscripted recordings 
for 2,336 hours (77.7\%), yielding an average unscripted-to-scripted ratio of approximately 
3.5:1. As shown in Table~\ref{tab:language_summary}, Kikuyu contributed the highest total at 
754 hours, followed by Dholuo (723 hrs), Kalenjin (521 hrs), Maasai (505 hrs), and Somali 
(502 hrs). Full transcription coverage for unscripted data ranged from 60\% for Dholuo (approximately 317 hours transcribed) to 100\% for both Somali and Maasai. Maasai and Somali exhibited 
the highest proportion of unscripted content, over 90\% of their respective totals, while 
Kikuyu and Dholuo maintained more balanced scripted/unscripted distributions. Scripted recordings averaged ~6 seconds, whereas unscripted recordings were longer and more variable, ranging from 45 to 65 seconds across languages.
The full breakdown of scripted and unscripted hours per language 
and domain is presented in Table~\ref{tab:topic_hours}.
\subsection{Demographic analysis}
\begin{figure}[!ht]
\begin{center}
\includegraphics[trim={0 0.5cm 0 0.7cm},clip,width=\columnwidth]
{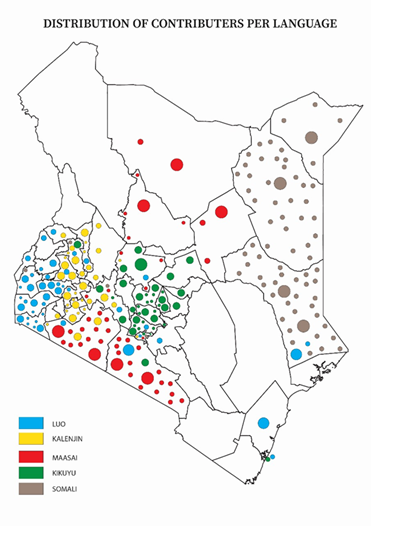}
\caption{Distribution of language contributors across counties in Kenya.}
\label{fig:contributors}
\end{center}
\vspace*{-2em}
\end{figure}

 Of the total 4,677 contributors, 3,381 (72.2\%) participated in Unscripted content, while 1,296 (28.7\%) contributed to Scripted recordings. Below we provide statistics across both modes:
 \paragraph{Geographical distribution: } There was great geographical distribution of the contributors with majority of them coming from the counties where the languages are predominately spoken and with a mix from urban regions like Nairobi. The geographical 
distribution of contributors across counties is illustrated in 
Figure~\ref{fig:contributors}.

 \paragraph{Gender:} In general, females accounted for 40.6\% of the contributors, while males 48.2\% and 11.1\% of unknown gender entries. Male contributors dominated most languages, particularly among Somali (60.0\%) and Maasai (54.2\%) speakers, whereas Kikuyu exhibited the strongest female representation at 54.4\%.
 
\paragraph{Age:} Contributors aged \textit{18–29} constitute the largest demographic, representing 39.3\% of the total dataset. The \textit{30–40} group accounts for 15.8\%, while those aged \textit{41–50} and \textit{51+} collectively represent 5.2\% . Notably, about 39.5\% did not disclose age, largely due to missing data in scripted records.

\paragraph{Highest education:} Tertiary-level contributors account for 64.0\%  of the total sample, underscoring a strong representation of higher-educated individuals. High School participants make up 20.1\% , Primary accounts for 2.6\% , and Unknown education levels comprise 12.0\%.

\paragraph{Employment status:} Students (25.7\%) and Unemployed contributors (27.8\%) dominate the dataset, indicating a significant proportion of participants who are either pursuing education or not currently engaged in employment. Self-employed individuals comprise 25.6\% , while Employed participants represent 8.8\% overall. The Unknown category accounts for 11.5\%  of all records.

\paragraph{Dialect:} Dialectal distribution within each language reveals varying levels of representation. In Dholuo, Milambo accounted for 49.3\% of the data, while Nyanduat constituted 50.7\%, reflecting a near-equal distribution. Kikuyu exhibited greater variation, with Gĩ-Kabete contributing 36.7\%, Ki-Murang'a 31.6\%, Ki-Mathira 23.5\%, and Kirinyaga 8.2\%. For Somali, all recordings were from the Maxatiri dialect. In Kalenjin, Kipsigis represented 56.7\% and Nandi 43.3\%, while in Maasai, the Maasai dialect predominated at 97.0\% compared to Samburu at 3.0\%. Unknown dialect entries accounted for 164 recordings overall.


\begin{table*}
\begin{center}
\small
\caption{Scripted (S) and Unscripted (UN) hours per language with contributor counts.}
\begin{tabular}{@{}lrrrr@{}}
\toprule
\textbf{Language} & \textbf{S (hrs)} & \textbf{UN (hrs)} & \textbf{Total} & 
\textbf{Contributors} \\
\midrule
Dholuo   & 195 & 528 & 723 & 1,195 \\
Kikuyu   & 183 & 571 & 754 & 1,331 \\
Kalenjin & 122 & 399 & 521 &   730 \\
Somali   & 118 & 384 & 502 &   633 \\
Maasai   &  51 & 454 & 505 &   788 \\
\midrule
\textbf{Total} & \textbf{669} & \textbf{2,336} & \textbf{3,005} & \textbf{4,677} \\
\bottomrule
\end{tabular}
\label{tab:language_summary}
\end{center}
\vspace*{-1em}
\end{table*}

\begin{table*}[!ht]
\centering
\small
\caption{Scripted (S) and Unscripted (UN) hours per language by topic (see 
Section~\ref{sec:domains} for domain descriptions).}
\label{tab:topic_hours}
\resizebox{\textwidth}{!}{%
\begin{tabular}{lrrrrrrrrrr}
\toprule
\textbf{Topic} & \multicolumn{2}{c}{\textbf{Dholuo}} & 
\multicolumn{2}{c}{\textbf{Kikuyu}} & 
\multicolumn{2}{c}{\textbf{Maasai}} & 
\multicolumn{2}{c}{\textbf{Kalenjin}} & 
\multicolumn{2}{c}{\textbf{Somali}} \\
\cmidrule(lr){2-3} \cmidrule(lr){4-5} \cmidrule(lr){6-7} 
\cmidrule(lr){8-9} \cmidrule(lr){10-11}
 & S & UN & S & UN & S & UN & S & UN & S & UN \\
\midrule
Agriculture \& Food        & 22.96 & 200.30 & 17.35 & 250.60 & 11.00 & 197.52 & 6.26  & 152.83 & 52.66 & 193.53 \\
Everyday Scenarios         & 42.40 & 41.68  & 21.83 & 54.21  & 12.99 & 59.79  & 31.32 & 42.11  & 14.26 & 44.39  \\
Financial Transactions     & 7.26  & 28.49  & 2.20  & 37.40  & --    & 33.14  & 0.02  & 36.34  & 1.26  & 21.73  \\
Digital Government Services& 7.03  & 23.29  & 1.04  & 25.72  & --    & 22.81  & 0.03  & 17.20  & 4.40  & 18.36  \\
Named Entity Recognition   & --    & 34.16  & 1.95  & 34.14  & --    & 22.00  & --    & 29.57  & --    & 23.27  \\
Role Play                  & --    & 27.73  & 0.75  & 27.66  & --    & 21.74  & 0.01  & 20.22  & --    & 11.06  \\
Extempore Stories          & --    & 32.99  & 16.01 & 28.27  & --    & 13.89  & --    & 14.22  & 1.50  & 12.27  \\
Healthcare                 & 17.35 & 42.70  & 17.48 & 31.77  & 12.02 & 28.68  & 26.44 & 32.13  & 20.29 & 24.86  \\
News \& Media              & 55.85 & 32.38  & 49.45 & 19.69  & 7.09  & 12.79  & 37.17 & 12.32  & 8.54  & 11.35  \\
Education \& Technology    & 1.34  & 37.45  & 14.19 & 37.71  & 1.51  & 26.19  & 6.51  & 26.57  & 5.45  & 17.22  \\
Customer Care Scenarios    & --    & 27.15  & 1.22  & 24.30  & 0.00  & 16.37  & --    & 16.17  & --    & 6.71   \\
General                    & 40.94 & --     & 39.58 & --     & 6.87  & --     & 15.21 & --     & 6.20  & --     \\
Climate Change             & --    & --     & --    & --     & --    & --     & --    & --     & 3.90  & --     \\
\bottomrule
\end{tabular}%
}
\vspace*{-1.5em}
\end{table*}
\subsection{Error Analysis}
This section examines rejection rates across languages, comparing scripted and unscripted datasets to identify patterns in data quality and collection conditions. It also presents an analysis of the primary factors contributing to data rejection in the collection process. 
Rejection rates exhibited variation across languages and between data collection modalities, with values ranging from approximately 1.5\% to 5\% for scripted recordings and from 6\% to 17\% for unscripted recordings. Scripted data exhibited high and relatively uniform quality due to predefined sentences that were read, which minimized variability and recording irregularities. In contrast, unscripted data showed greater variability, with more disfluencies, code-switching, inconsistent speaking pace, noise overlap and truncated responses, leading to higher rejection rates.

Rejection reasons were analyzed based on validator descriptions, with representative examples provided for each sub-category (see Table \ref{tab:rejection-categories-examples}). Data quality issues overwhelmingly dominated, accounting for 99.9\% of rejected submissions. General poor recording quality was the leading cause (51\%), reflecting recordings with overall distortion or unintelligibility. Background noise accounted for 30.5\%, underscoring the impact of recording environments. Additional factors included excessive pauses or slow speech (9.4\%) and unclear or low-volume speech (9\%). These findings highlight the need for pre-submission quality checks, standardized rejection categories, and improved user feedback mechanisms to minimize avoidable data loss.

\begin{table*}[]
	\centering
	\caption{Distribution of rejection reasons by category and sub-category}
	\label{tab:rejection-categories-examples}
    \resizebox{\textwidth}{!}{%
	\begin{tabular}{p{3.2cm} p{4.2cm} p{2.2cm} p{5.8cm}}
		\hline
		\textbf{Main Category} & \textbf{Sub-category} & \textbf{Percentage (\%)} & \textbf{Example Annotations} \\
		\hline
		Data Quality Issues 
		& General poor recording quality 
		& 51 
		& poor quality record; bad record; not acceptable; technical problem; problematic \\
		& Background noise 
		& 30.5 
		& background noise encountered; noisy environment; too much background sound \\
		& Excessive pauses / slow speech 
		& 9.4 
		& poor quality due to too much pauses; slow reading and more pauses \\
		& Unclear / low-volume speech 
		& 9 
		& voice record is unclear; relatively low voice; low voice not audible \\
		\hline
		Annotation \& Linguistic Issues 
		& Language mismatch 
		& 0.06 
		& irrelevant record, with more English than expected language \\
		\hline
		Technical \& Compliance Issues 
		& Format / metadata problems 
		& 0.04 
		& file format issue; metadata problem \\
		\hline
	\end{tabular}}
    \vspace*{-1.5em}
\end{table*}
\section{Ethical Considerations}
\label{sec:ethics}
 
Ethical practices, including informed consent, participant anonymity, and cultural sensitivity, 
guided all activities throughout the project.
 \vspace*{-0.5em}
\subsection{Ethical Approval and Consent}
\label{sec:consent}
The AfriVoices-KE project received formal approval from the host institution Review Board and the National level
research permit, ensuring compliance. All contributors 
provided informed consent before accessing the data collection platform, with detailed explanations of data usage, storage protocols, and participant rights. During on-boarding, participants were briefed on the project and given the opportunity to ask questions and seek clarification. Contributors were required to read the project terms and provide consent before creating a profile in the data collection app. Identifiable information like phone number and ID were used solely for payment processing and not disclosed. Voice recordings were anonymized using unique identification numbers that disconnected personal 
identifiers from audio files, and stored securely using encrypted cloud infrastructure with role-based access controls.
\vspace{-0.25em}
\subsection{Cultural Considerations}
\vspace{-0.3em}
\label{sec:cultural}
Cultural sensitivity was prioritized in prompt design and participant engagement, with input from 
community leaders and native speakers ensuring community-centered approaches throughout. Native speaker consultants guided prompt 
design and validation to address cultural nuances, e.g. taboos in Healthcare discussions. Dialects prevalent in formal media were prioritized for broader intelligibility: 
for Dholuo, the standard dialect used in Luo radio stations was emphasized; for Kalenjin, 
sub-dialects Kipsigis and Nandi, common in agricultural and community contexts, were included; 
Maasai recordings incorporated both Maasai and Samburu varieties used in storytelling; and Kenya 
Somali prioritized dialects from news broadcasts.  

\subsection{Compensation}
\label{sec:compensation}
The project compensated all roles involved in data collection and management at rates set above 
Kenya's minimum wage.\footnote{An exchange rate of approximately Ksh~130 per USD is used 
throughout.} For scripted collection, translators and sentence generators received Ksh~10 (USD~0.08) per 
sentence. Prompt contributors received Ksh~10 (USD~0.08) per text prompt and Ksh~15 (USD~0.12) per image prompt, each including a local language version and English translation.
Voice contributors received Ksh~2,000 (USD~16) per validated hour of recordings, and voice validators for unscripted collection were paid Ksh~3 per validated sentence-voice pair. Transcribers were 
paid Ksh~5,000 (USD~38) per hour, with transcription reviewers compensated at Ksh~3,500 (USD~27) per hour of review.
Field mobilizers and resource persons received daily allowances covering airtime and internet bundles (Ksh~1,000), accommodation (Ksh~3,000--5,000 per night), venue hire (Ksh~3,000--5,000 per day), meals (Ksh~1,000 per meal), and participant refreshments (Ksh~200 per participant), with transport reimbursed according to region. 
For languages with dedicated recruitment coordinators, such as Maasai and Somali, coordinators received Ksh~700 per participant who was successfully on-boarded and completed the required one hour of validated voice.

\section{Dataset Release and Use}
\label{sec:release}
The AfriVoices-KE dataset is publicly available on Hugging 
Face\footnote{\url{https://huggingface.co/Anv-ke}} under a CC BY 4.0 license, with access managed through a request 
form to track usage. The dataset is continuously updated, and users are advised to cite the 
latest version in publications and benchmarking work. Standard machine learning splits are 
provided, train (85\%), dev (5\%), dev-test (5\%), and a hidden test (5\%), with speaker-disjoint sets 
to ensure fair evaluation. A strict ethical disclaimer prohibits use for surveillance, 
discrimination, or exploitation.

The dataset card provides comprehensive documentation covering dataset composition across the 
five languages described in Section~\ref{sec:languages}, domain distribution as detailed in 
Table~\ref{tab:topic_hours}, demographic information, and transcription guidelines. Audio files 
are stored with metadata specifying language, dialect, domain, speaker information, and 
recording type (scripted or unscripted), enabling researchers to conduct stratified analyses and 
develop language-specific models. The resource is designed to support training of automatic 
speech recognition systems, low-resource language research, and benchmarking of multilingual 
speech technologies. All datasets and documentation are maintained in open repositories to support ongoing community contributions and improvements.

 \section{Acknowledgements}
\label{sec:ack}
The authors gratefully acknowledge the contributions of the project staff, language leads, resource persons, respondents, annotators, and linguists, whose collective expertise and dedication were instrumental in the conceptualization, data collection, and successful implementation of this project. 
We further extend our sincere appreciation to our collaborating partners—University of Pretoria, Way With Words, Data Science Nigeria, and Digital Umuganda—for their invaluable technical input and collaborative support throughout the project lifecycle. 
We also acknowledge the insightful contributions of colleagues from Google and Meta Platforms, whose expertise and engagement significantly enriched the design and execution of this work.

This work was supported, in whole, by the Gates Foundation, Investment No: INV-074419. Under the grant
conditions of the Foundation, a Creative Commons
Attribution 4.0 Generic License has already been
assigned to the Author Accepted Manuscript version
that might arise from this submission.

\section{References}

\bibliographystyle{lrec2026-natbib}
\bibliography{lrec2026-example}

\label{lr:ref}
\bibliographystylelanguageresource{lrec2026-natbib}
\bibliographylanguageresource{languageresource}

\end{document}